\documentclass[10pt, journal]{IEEEtran}
\pdfoutput=1
\usepackage[final]{graphicx}
\usepackage{amsmath}
\usepackage{subfig}
\usepackage{float}
\usepackage{array}
\usepackage{mathrsfs}
\newcolumntype{C}[1]{>{\centering\arraybackslash}p{#1}}
\usepackage{multirow}
\usepackage{booktabs}
\usepackage{amssymb,times}
\usepackage{cite}
\usepackage[ruled,vlined]{algorithm2e}     

\IEEEoverridecommandlockouts

\begin{document}

\title{Mask-RadarNet: Enhancing Transformer With Spatial-Temporal Semantic Context for Radar Object Detection in Autonomous Driving}

\author{Yuzhi Wu, Jun~Liu,~\emph{Senior Member, IEEE}, Guangfeng Jiang, Weijian~Liu,~\emph{Senior Member, IEEE}, and~Danilo~Orlando,~\emph{Senior Member, IEEE} 
\thanks{Yuzhi Wu, Jun Liu and Guangfeng Jiang are with the Department of Electronic Engineering and Information Science, University of Science and Technology of China, Hefei 230027, China (e-mail: yuzhiwu1105@mail.ustc.edu.cn; junliu@ustc.edu.cn; jgf1998@mail.ustc.edu.cn).

Weijian Liu is with Wuhan Electronic Information Institute, Wuhan 430019, China (e-mail: liuvjian@163.com).

Danilo Orlando is with the Universit\`{a} degli Studi ``Niccol\`{o} Cusano'', 00166 Rome, Italy (e-mail: danilor78@gmail.com).
}}

\maketitle

\begin{abstract}
As a cost-effective and robust technology, automotive radar has seen steady improvement during the last years, making it an appealing complement to commonly used sensors like camera and LiDAR in autonomous driving. Radio frequency data with rich semantic information are attracting more and more attention. Most current radar-based models take radio frequency image sequences as the input. However, these models heavily rely on convolutional neural networks and leave out the spatial-temporal semantic context during the encoding stage.
To solve these problems, we propose a model called Mask-RadarNet to fully utilize the hierarchical semantic features from the input radar data. Mask-RadarNet exploits the combination of interleaved convolution and attention operations to replace the traditional architecture in transformer-based models. In addition, patch shift is introduced to the Mask-RadarNet for efficient spatial-temporal feature learning. By shifting part of patches with a specific mosaic pattern in the temporal dimension, Mask-RadarNet achieves competitive performance while reducing the computational burden of the spatial-temporal modeling. In order to capture the spatial-temporal semantic contextual information, we design the class masking attention module (CMAM) in our encoder. Moreover, a lightweight auxiliary decoder is added to our model to aggregate prior maps generated from the CMAM. Experiments on the CRUW dataset demonstrate the superiority of the proposed method to some state-of-the-art radar-based object detection algorithms. 
With relatively lower computational complexity and fewer parameters, the proposed Mask-RadarNet achieves higher recognition accuracy for object detection in autonomous driving.
\end{abstract}

\begin{figure}[htbp]
\centering
\includegraphics[scale=0.4]{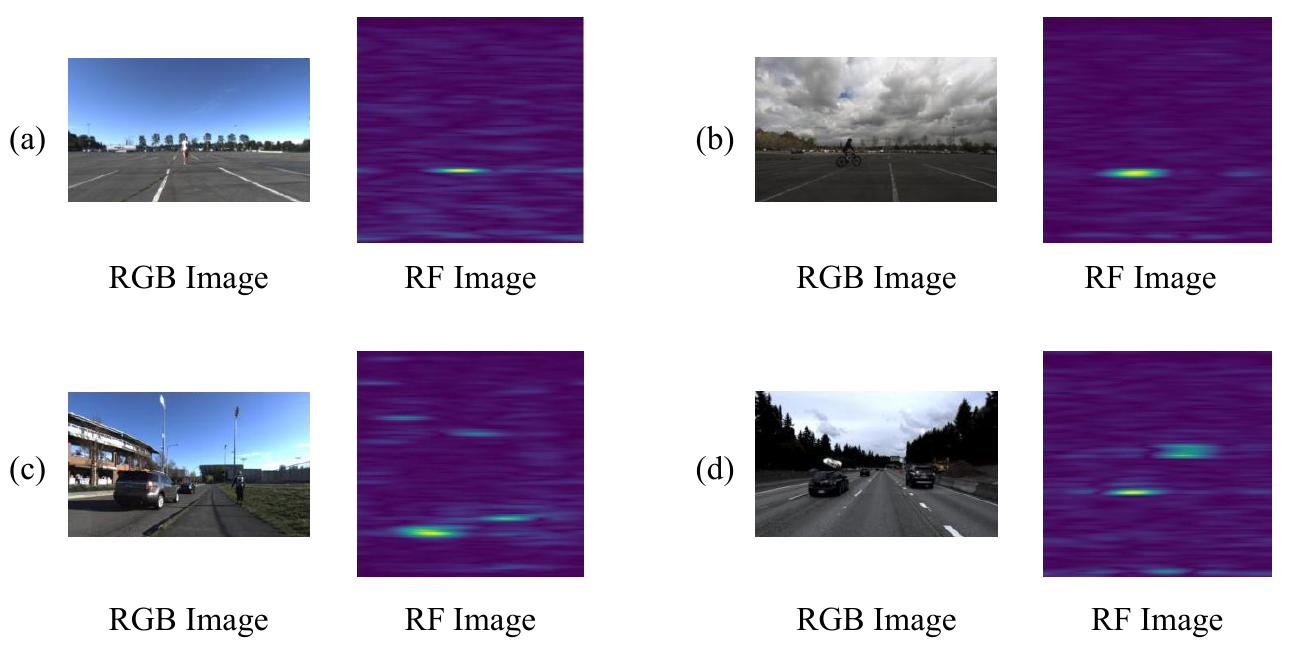}
\caption{Examples of RGB images and their corresponding RF images which represent the same scene. RF images are in range-azimuth coordinates.}
\label{figure1}
\end{figure}

\IEEEpeerreviewmaketitle
\section{Introduction}
\IEEEPARstart{}{}
\IEEEPARstart{L}{ast} decades have witnessed growing interests in autonomous driving \cite{10196355,9351818,9284628,5940562,8575295,10265760}, and object detection is one of the most fundamental task for practical deployment. To better capture the surrounding objects, sensors equipped in autonomous cars are receiving increasing attention. Among the commonly used sensors, millimeter-wave (MMW) frequency-modulated continuous-wave (FMCW) radar has the following unique advantages: 1) the narrow MMW band allows radar signal to penetrate through fog and smoke, which is crucial in extreme weather conditions; 2) FMCW radar has better acquisition capabilities for detecting longer ranges; 3) FMCW radar is robust to lighting while being cheap. 
However, because of the difficulties in deciphering significant clues for semantic understanding, radar is frequently regarded as a complement sensor for RGB cameras and LiDARs.
Comparatively, the RGB images and point cloud data from cameras and LiDARs are relatively easy for human to understand since the semantic information they convey is obvious \cite{9523092}. For example, Fig. \ref{figure1} shows some RGB images and their corresponding radio frequency (RF) images which represent the same scene. In recent work\cite{9523163,8803392,8904867}, FMCW radar is merely processed to provide location and speed information for the detected objects without fully exploiting the semantic information. In other words, the development of object detection with FMCW radar is still in its early stages, making it worthwhile to explore further.

Radar data are usually represented in two different formats, i.e., RF images and radar point clouds. Considering that the current 3D radar point clouds are too sparse to detect objects accurately \cite{8455344,9157505,9353210}, many researchers start to take advantage of RF images\cite{9022248,9150751}. In the field of traditional signal processing\cite{4815550}, peak detection algorithms, such as those with a constant false alarm rate, are utilized in RF images to determine the object's location. Subsequently, a classifier is employed to identify the object's category\cite{Angelov2018PracticalCO}. With the emergence of deep learning, the focus of research is naturally shifted to extract RF image features via neural networks. It has to be pointed out that many labeled data are required for training neural networks. However, it is more difficult to annotate RF images than RGB images due to the abstract semantic information, especially for object detection task. Zhang \emph{et al}.\cite{9469418} proposed an instance-wise auto-annotation method to build a new radar dataset called RADDet. But the size of this dataset is small. Ouaknine \emph{et al}.\cite{9413181} presented a semi-automatic annotation approach and proposed a dataset of synchronized camera and radar recordings with annotations. Recently, Wang \emph{et al}.\cite{9353210} developed a cross-modal supervision framework to annotate object labels on RF images automatically with a camera-radar fusion (CRF) strategy, and built a new benchmark for radar object detection task. During the training stage, annotations of object are processed into confidence maps as the ground truth. To test the models, the output is post-processed like\cite{1699659} to generate final results.

With the open access of radar RF image datasets, much work \cite{9353210,10.1145/3460426.3463657,10.1145/3460426.3463654} uses 3D convolutional neural network (CNN) to extract semantic and velocity information from multi-frame RF images. While these models perform well in certain tasks, the extensive 3D convolutions come with a high computational cost, which may be inappropriate for real-time applications. Besides, 3D CNN shows poor performance in extracting global features and it cannot acquire the dependencies between multi-frame radar RF images well. The work in \cite{9989400} was the first one to introduce the transformer-based model into radar object detection. The model is a U-shaped one containing convolution and attention operations. Although the architecture facilitates the extraction of multiscale features, it overlooks the significance of the spatial-temporal semantic context of the attention maps, leading to some misclassified results.

To solve the issues mentioned above, we propose a novel model called Mask-RadarNet, a 3D transformer for radar object detection. The Mask-RadarNet exploits the combination of interleaved convolution and self-attention operations. The hybrid architecture enables the encoder of Mask-RadarNet to extract local and global features effectively. We utilize a simple but effective method called patch shift \cite{DBLP:journals/corr/abs-2207-13259} for efficient spatial-temporal modeling in the 3D transformer. This attempt enhances spatial-temporal feature learning efficiently for our model. Moreover, we design a class masking attention module (CMAM) in our encoder to capture the spatial-temporal contextual information. Although RF images are non-intuitive compared with RGB images, and much more difficult for human eyes to understand, we still hold the belief that the spatial-temporal semantic context contained in RF image sequences is crucial for radar object detection. With the supplement of the spatial-temporal semantic context, the CMAM enhances the global feature acquired by attention operations. It also generates prior maps for our task and guides the model to update during the training stage. Besides, we add a semantic auxiliary decoder to aggregate prior maps from different stages. As shown in Fig. \ref{figure2}, our model achieves the state-of-the-art (SOTA) performance on the CRUW dataset. In brief, our work has made the following contributions:

\begin{figure}[htbp]
\centering\includegraphics[scale=0.58]{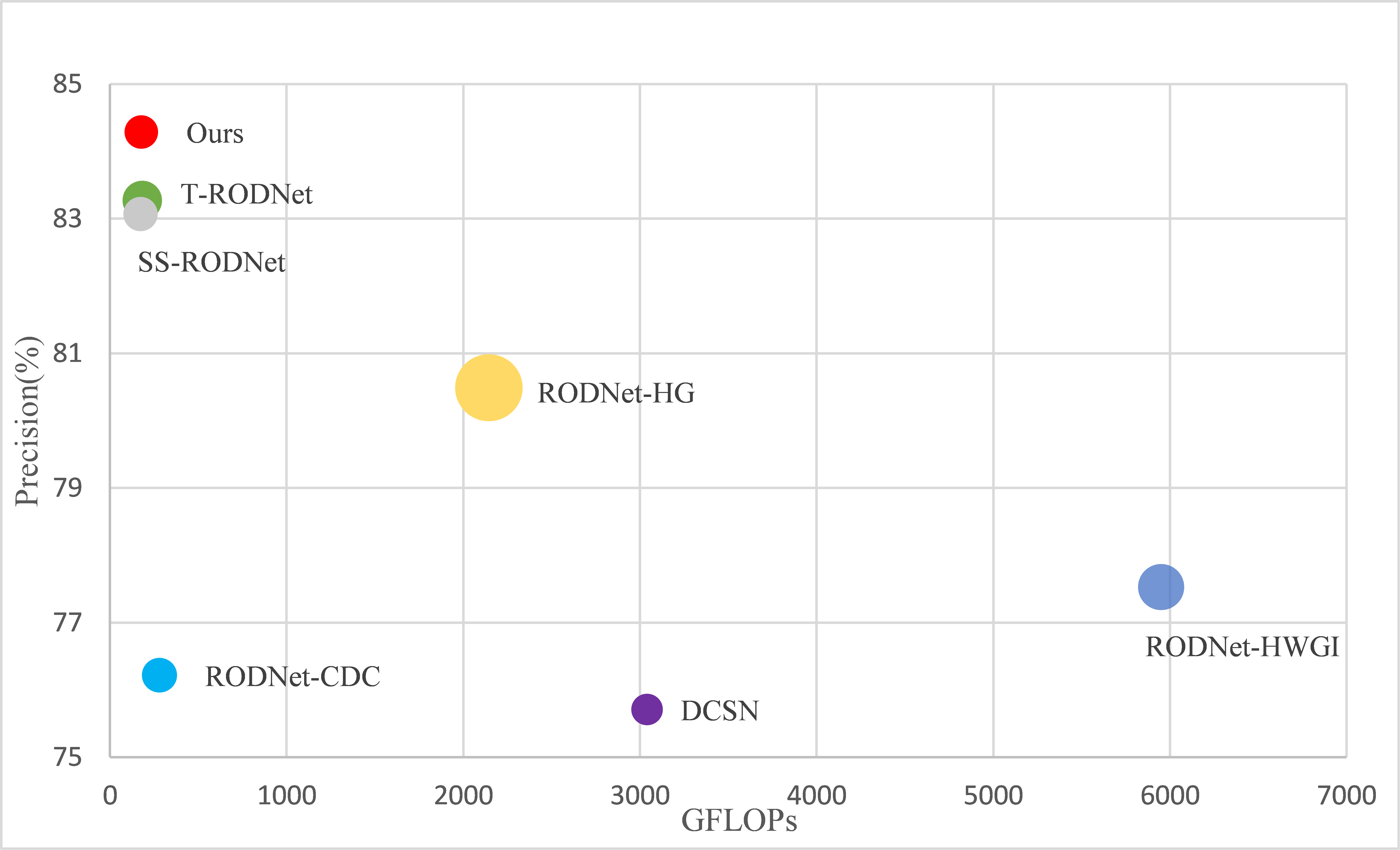}
\caption{Comparisons of Mask-RadarNet with other SOTA models on the CRUW dataset. Different models are represented by marks of different colors. Moreover, a smaller mark means a smaller
model size. }
\label{figure2}
\end{figure}
1) We propose Mask-RadarNet that is a 3D Transformer combining interleaved convolution and self-attention operations for radar object detection. The proposed Mask-RadarNet achieves significant improvements in object detection performance over previous models on the CRUW dataset.

2) We introduce patch shift in the Mask-RadarNet for efficient spatial-temporal feature learning. Our model achieves competitive performance with other methods while reducing the computational burden of spatial-temporal modeling. 

3) We design a specific module called CMAM to capture the spatial-temporal contextual information and enhance the global feature with spatial-temporal semantic context. Besides, we add an auxiliary decoder to generate prior maps during the training stage. 

The rest of this article is structured as follows. In Section II, we introduce some related work. Section III describes the design of Mask-RadarNet in detail. The experimental results and ablation studies are included in Section IV. Section V concludes with conclusions.

\section{Related Work}
In Sections II-A, we discuss the current radar-based perception methods for autonomous driving. Then we review some transformer-based methods for various computer vision tasks in Sections II-B. Besides, we introduce some work about semantic context for various computer vision tasks in Sections II-C.

\subsection{Radar-Based Perception Methods for Autonomous Driving}
There are mainly two representations of radar data: one is the dense raw RF images, and the other is the sparse radar point clouds. In Sections II-A, we review related work on perception for autonomous driving from the perspective of two different formats.
\subsubsection{RF image}
With the advancement in deep learning, a series of research explores neural networks to extract features from raw RF images. Capobianco \emph{et al}. \cite{10.1007/978-3-319-75608-0_9} employed CNN to recognize vehicle categories from range-doppler images.
Angelov \emph{et al}.\cite{Angelov2018PracticalCO} considered a modular pipelined framework on raw radar data, and explored three distinct kinds of neural networks including convolution-based ones to classify radar objects. Gao \emph{et al}. \cite{8458209} used a modified complex-valued convolutional neural network to enhance radar imaging. Zhang \emph{et al}. \cite{9469418} utilized a residual network as the backbone and proposed a dual detection head for more accurate predictions. In order to extract semantic and velocity information from multi-frame radar images, some work was proposed to use 3D convolution. Hazara \emph{et al}. \cite{8821302} employed a model based on 3D CNN to acquire embedding features from radar data with a distance-based triplet-loss similarity metric.
Wang \emph{et al}. \cite{9353210} proposed a stacked-hourglass model on multiframe RF images to generate predictions. Hsu \emph{et al}. \cite{10.1145/3460426.3463657} further adopted dilated convolution in the backbone network to achieve a larger receptive field. The work in \cite{10.1145/3460426.3463654} used the squeeze-and-excitation network to predict the location and category of the object. RaLiBEV \cite{bu6} proposed a novel fusion paradigm with radar range-azimuth heatmaps and LiDAR point clouds, and designed an anchor-free detector based on the fused features. Inspired by DETR \cite{bu9}, LQCANet \cite{bu7} also employed the learnable-query for feature fusion at various scales. T-RODNet \cite{9989400} introduced a transformer-based model in radar object detection that contains convolutions and attention operations, with the intention of utilizing the ability of both to acquire local and global features simultaneously. SS-RODNet \cite{bu8} further proposed a lightweight model by pretraining radar spatial-temporal information. 
\subsubsection{Radar Point Cloud}
In the current real-world automotive application, radar suppliers commonly provide radar point clouds for autonomous driving. As a lightweight data representation, point clouds provide an intuitive spatial structure of the surroundings. Liu \emph{et al}. \cite{bu1} believed that radar points with diverse semantic information rarely belong to the same object, and designed a clustering method based on semantic segmentation. Xiong \emph{et al}. \cite{bu2} proposed a contrastive learning method to address the problem of insufficient annotation of radar points, and designed a model that performs well with limited labeled radar points. Kernel density estimation branch is added to the pillar-based backbone for feature encoding in SMURF\cite{bu3}, alleviating the impact of sparsity in radar point clouds. Some work attempts to integrate radar point clouds and corresponding RGB images.
RCFusion \cite{bu4} utilized orthographic feature transform for transforming the image perspective view (PV) features into the bird’s-eye-view (BEV) domain, and then fused image BEV features and radar BEV features using interactive attention module. LXL \cite{bu5} generated radar occupancy grids and predicted image depth distribution maps separately, which both assist in converting image PV features to BEV features, so that the image features can be aligned with radar BEV features. Although radar point clouds have advantages in being a lightweight data representation, they suffer from the inevitable loss of potential information in raw radar tensors during signal processing \cite{9022248}, which may cause the failure in detecting small objects.

\subsection{Transformer-based Methods for Various Computer Vision Tasks}
After being developed in the field of natural language processing (NLP) \cite{2017Attention},
transformer-based methods have gained popularity for various computer vision tasks following vision transformer (ViT)\cite{Dosovitskiy2020AnII}. The transformer-based methods have produced outstanding results on semantic segmentation \cite{10254444,Wang2020MaXDeepLabEP}, object detection \cite{bu9,Zhu2020DeformableDD}, image generation\cite{chang2022maskgit,esser2020taming}, video segmentation \cite{Bertasius2021IsSA,liu2021video} and other computer vision tasks\cite{Pu2022EDTERED,Wang2021TransBTSMB,Xie2022DXMTransFuseUD}. Liu \emph{et al}.\cite{9710580} adopted the shifted window-based approach in ViT architectures, which greatly enhances performance. Other work followed this approach.
Cao \emph{et al}.\cite{Cao2021SwinUnetUP} built a U-shape transformer-based model that employs the hierarchical transformer architecture with shifted windows as the backbone for feature extraction.
To further improve the semantic segmentation quality of medical images, Lin \emph{et al}.\cite{9785614} attempted to simultaneously integrate the advantages of the hierarchical transformer architecture with shifted windows into both encoder and decoder of the U-shaped model.

However, in comparison to CNNs, vision transformers still experience the drawback of image-specific inductive bias, leading to inefficiency in extracting local information. Some researchers try to investigate ways to improve the local feature modeling capabilities of ViTs. To model the relationships between tokens at different scales, Xu \emph{et al}.\cite{9710209} adopted a hybrid architecture that contains depthwise convolutions and cross-attention operations. Chu \emph{et al}.\cite{Chu2021TwinsRT} built a model upon pyramid vision transformer \cite{9711179} by combining depthwise separable convolutions and relative position embedding. Tu \emph{et al}.\cite{Tu2022MaxViTMV} proposed a model which involves MBConv-based convolution followed by block-wise attention operations and grid-wise attention operations. 

\subsection{Semantic Context for Various Computer Vision Tasks}
Computer vision tasks require semantic context information to get high-quality results. 
Chen \emph{et al}.\cite{7913730} proposed a module which employs atrous convolution to efficiently broaden the field of view of filters in order to include more context information. They further augmented this module with global average pooling in \cite{Chen2017RethinkingAC}. Zhao \emph{et al}. \cite{8100143} proposed a pyramid pooling module which can aggregate context from different region to leverage global context information. Yu \emph{et al}. \cite{8578297} added the global pooling on the top of the U-shape model with the purpose of encoding the global context. Zhang \emph{et al}. \cite{Zhang2018ContextEF} designed a new context encoding module that, by introducing prior information, improves the model's performance.
Jain \emph{et al}. \cite{Jain2021SeMaskSM} incorporated the semantic context of RGB images into the backbone of a hierarchical transformer model.
Zhang \emph{et al}. \cite{2018arXiv180308904Z} explored the impact of global contextual information in semantic segmentation. Jin \emph{et al}.\cite{Jin2021ISNetII} advocated enhancing pixel representations by combining the image-level and semantic-level contextual information.

\begin{figure*}[htbp]
\centering
\includegraphics[scale=0.48]{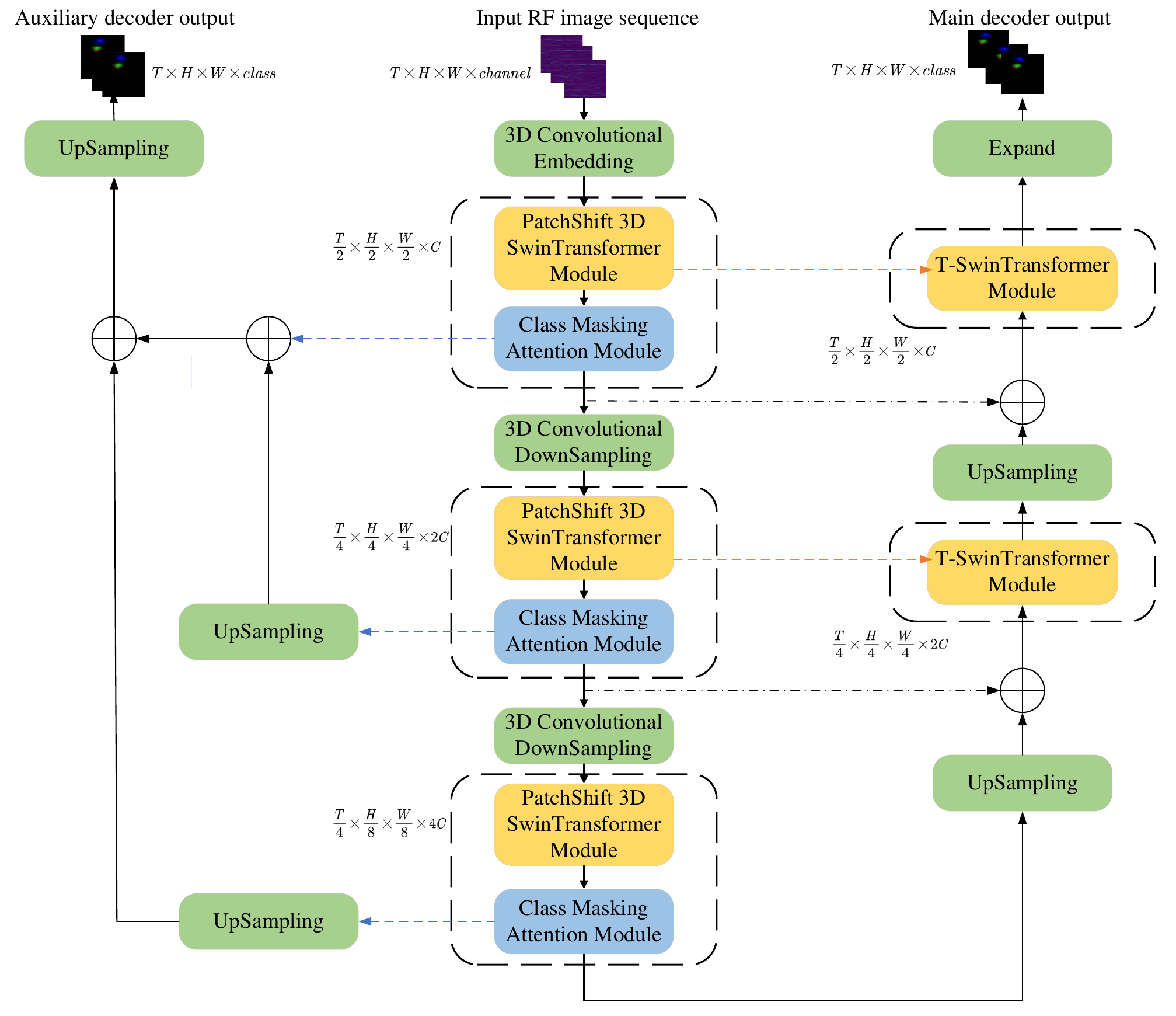}
\caption{Overview of proposed Mask-RadarNet. The encoder is in the middle, and the two decoders are on the left and right. The encoder is a hierarchical hybrid structure of convolution and self-attention mechanisms, which consists of the PatchShift 3D SwinTransformer module and the CMAM. The right decoder is the main decoder including the T-SwinTransformer module. The left decoder is the auxiliary decoder which generates the final prior maps. The orange lines represent the movement of query and key features from the PatchShift 3D SwinTransformer module to the main decoder. The blue lines represent the movement of query features from the CMAM to the auxiliary decoder.}
\label{figure3}
\end{figure*}

\section{Methodology}
\subsection{Overall}
The overall architecture of our Mask-RadarNet is presented in Fig. 3, which maintains one encoder and two different decoders: main decoder and semantic auxiliary decoder. Specifically, the encoder is a hierarchical hybrid structure of convolution and self-attention mechanisms for exploiting both local and global representations. We specially design the PatchShift 3D SwinTransfomer module and the CMAM to model and capture the spatial-temporal contextual information. Symmetrically, the main decoder involves two T-SwinTransformer modules, two upsampling layers and the last expanding layer for mask predictions. It not only adds skip connections between corresponding feature pyramids of the encoder and decoder, but also utilizes cross attention to fuse the features from the encoder and the inherent features from the decoder. Besides, we use a lightweight semantic auxiliary decoder during training to generate prior maps. The network architecture will be described in detail in the following sections.

\subsection{Encoder}
\subsubsection{PatchShift 3D SwinTransformer Module}
\begin{figure}[tbp]
\centering
\includegraphics[scale=0.3]{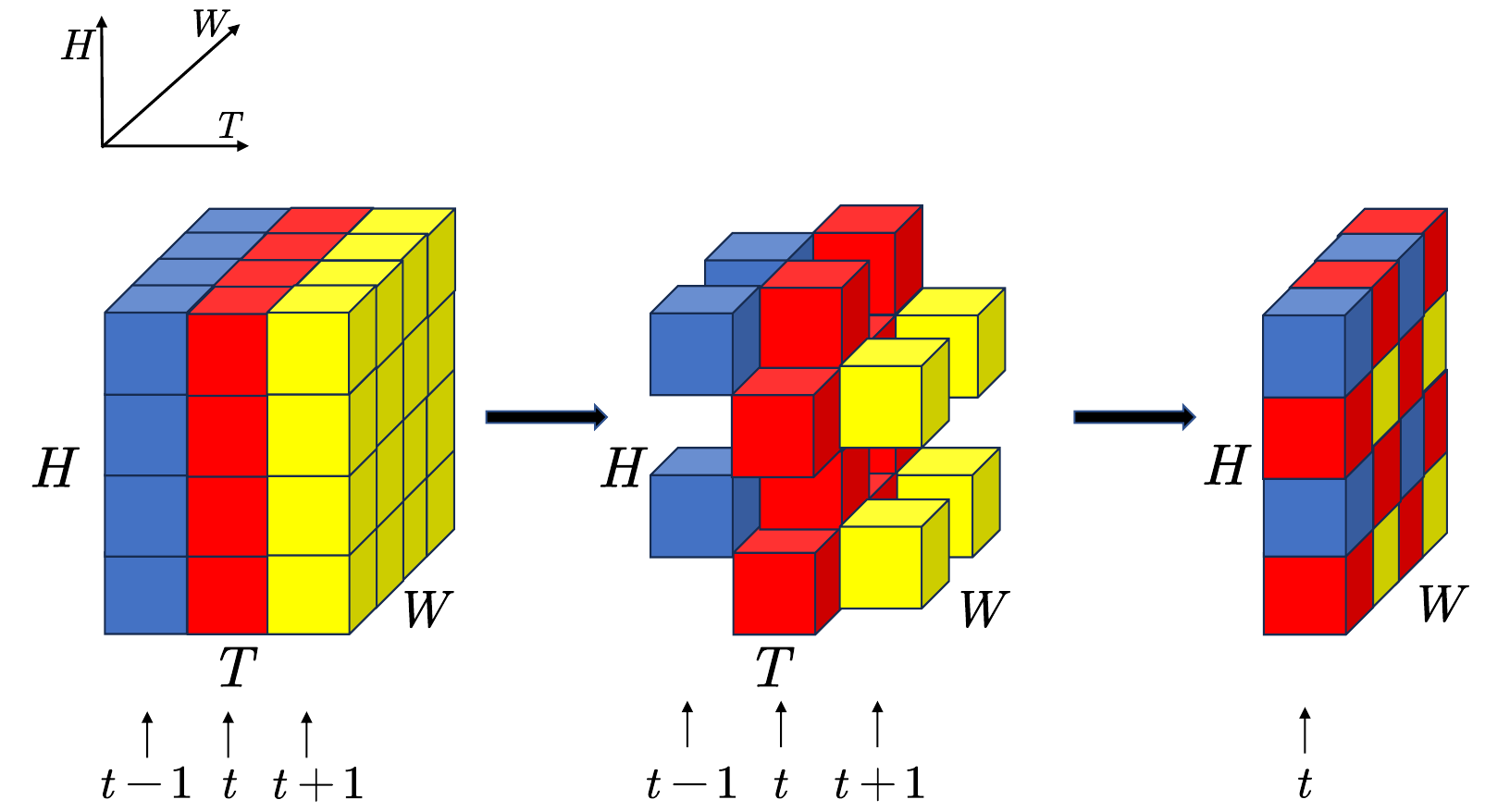}
\caption{An example of patch shift for three neighboring frames. The current frame $t$ aggregates information from neighboring frames $t-1$ and $t+1$.}
\label{figure4}
\end{figure}

\begin{figure}[tbp]
\centering
\includegraphics[scale=0.33]{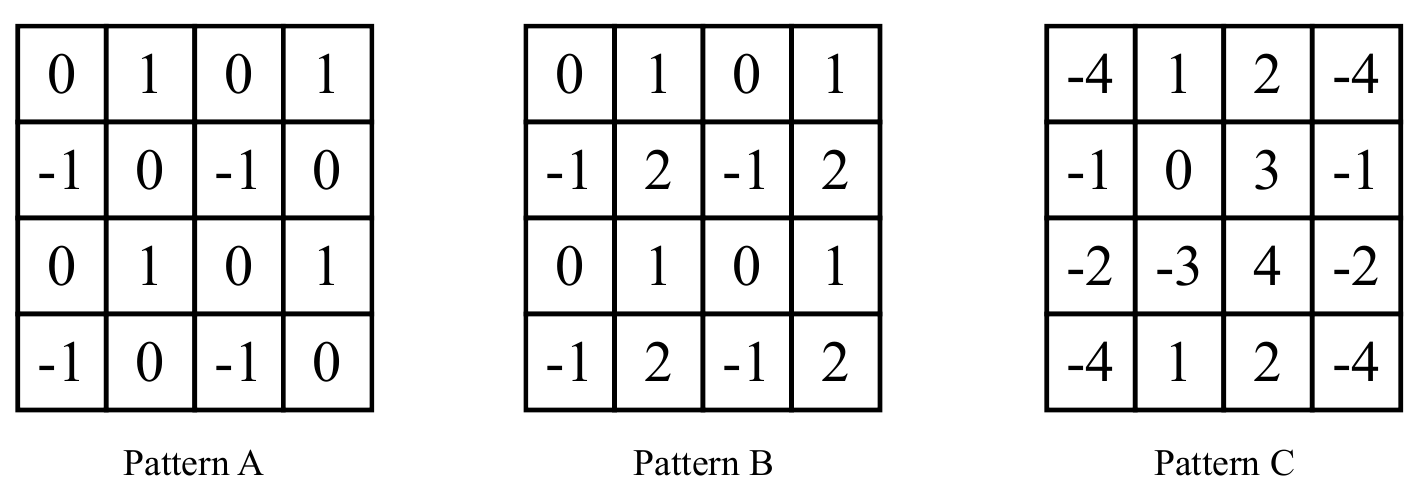}
\caption{Three typical shift patterns. Pattern A only shifts patches within 3 neighboring frames, while Pattern B has a temporal of 4 and Pattern C has a temporal field of 9.}
\label{figure5}
\end{figure}

Considering that the inter-frame details within RF image sequences play an indispensable role in assisting the network to accurately recognize targets, we propose PatchShift 3D SwinTransformer Module for spatial-temporal feature extraction. Transformer model with patch shift operation was first proposed by \cite{DBLP:journals/corr/abs-2207-13259} for action recognition. It was originally designed for RGB image sequences. Our model for the first time introduces it in RF image sequences for efficient spatial-temporal fusion. Generally, patch shift is an effective way for temporal modeling, which shifts the patches of input features along the temporal dimension following specific patterns. Fig. \ref{figure4} shows an example of patch shift for three neighboring frames, where the symbols $H$, $W$, and $T$ denote the height, width, and temporal dimension, respectively, and the blue, red and yellow colors represent the frames $t-1$, $t$, and $t+1$, respectively. Part of patches in red frame are replaced by patches from blue and yellow frames. This means the current frame $t$ aggregates information from other frames $t-1$ and $t+1$ with a specific pattern. Patch shift can be carried out with different patterns. Fig. \ref{figure5} depicts some typical shift patterns. The numbers denote the frame indices from which the patches are taken. The symbols ``0'', ``-'', ``+'' represent the current, previous, and next frames, respectively. To cover all patches, we continually apply shift pattern in a sliding window way \cite{9710580}. After patch shift operation, the spatial features mingle with the temporal feature in a zero-computation way. Hence we can directly exploit volume-based 3D transformer module.  

\begin{figure}[tbp]
\centering
\includegraphics[scale=0.285]{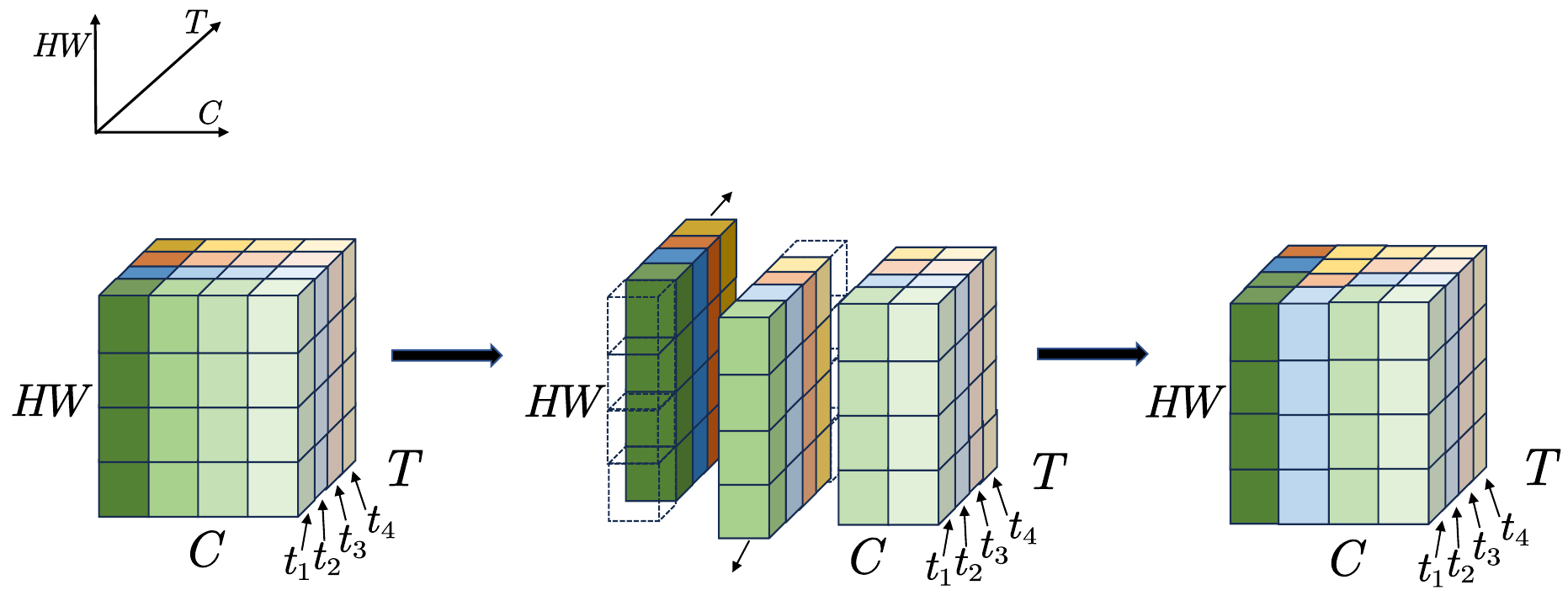}
\caption{An example of channel shift for four neighbouring frames. The first channels of frames $t_2$, $t_3$ and $t_4$ are replaced by those of frames $t_1$, $t_2$ and $t_3$. The second channels of frames $t_1$, $t_2$ and $t_3$ are replaced by those of $t_2$, $t_3$ and $t_4$. The rest remains unchanged.}
\label{figure6}
\end{figure}

\begin{figure}[htbp]
\centering
\includegraphics[scale=0.29]{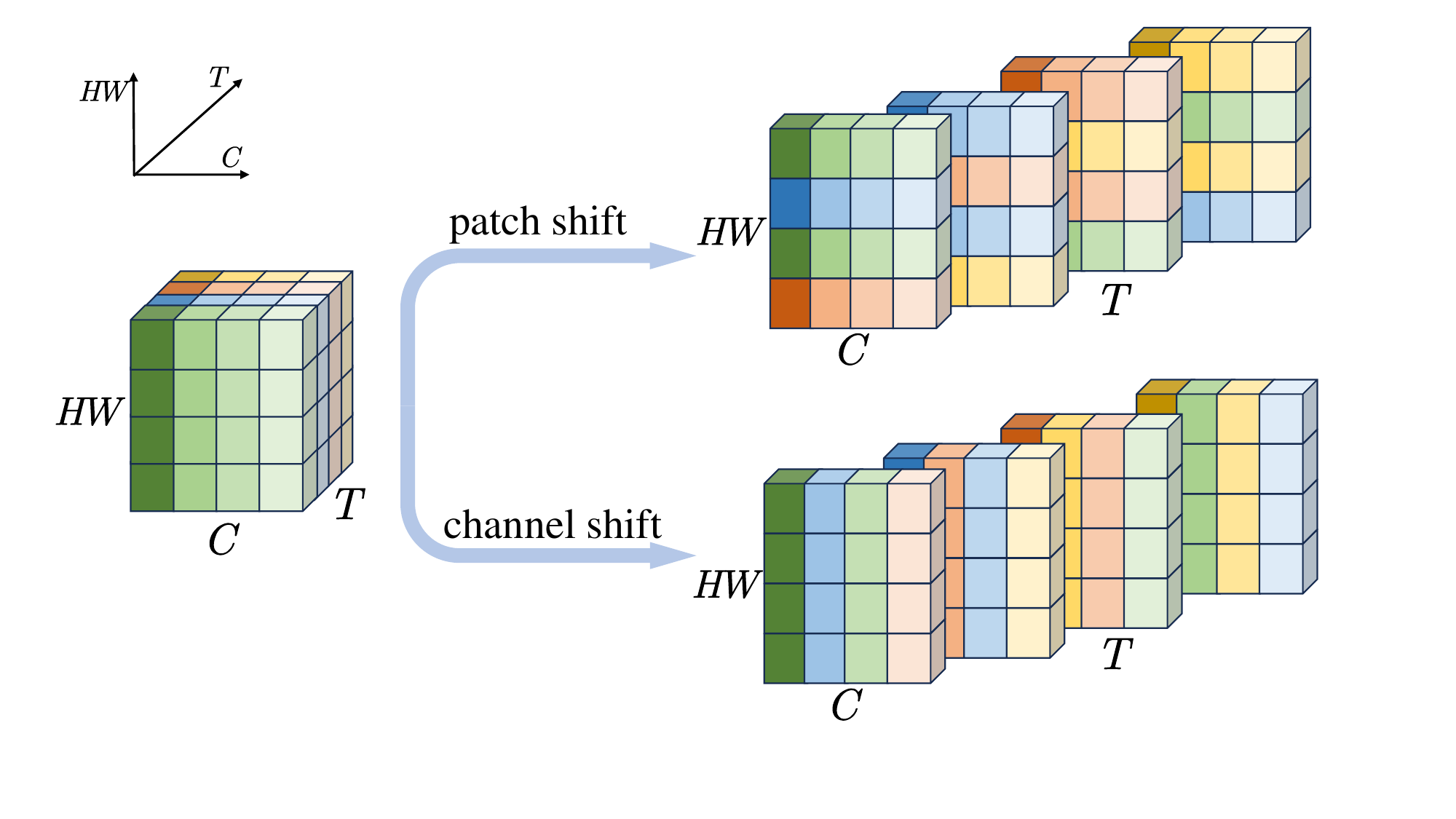}
\caption{Illustration of patch shift and channel shift. We can see that they perform shifting operations in orthogonal direction.}
\label{figure7}
\end{figure}

Patch shift operation learns temporal information by moving part of patches from other frames, thus keeping the full channel information of each patch. This means that patch shift is sparse in the spatial domain but dense in the channel domain. Channel shift is also a method for temporal modeling, which is just the opposite to patch shift. It replaces a constant proportion of channels in the current frame with other frames along the temporal dimension. As shown in Fig. \ref{figure6}, part of channels are shifted forward by one frame, while another part of channels are shifted backward by one frame, with the rest remaining unchanged. Fig. \ref{figure7} illustrates the differences between patch shift and channel shift. We describe a tensor with flattened spatial dimension $HW$, temporal dimension $T$ and channel dimension $C$. 
\begin{figure}[!h]
\centering
\includegraphics[scale=0.41]{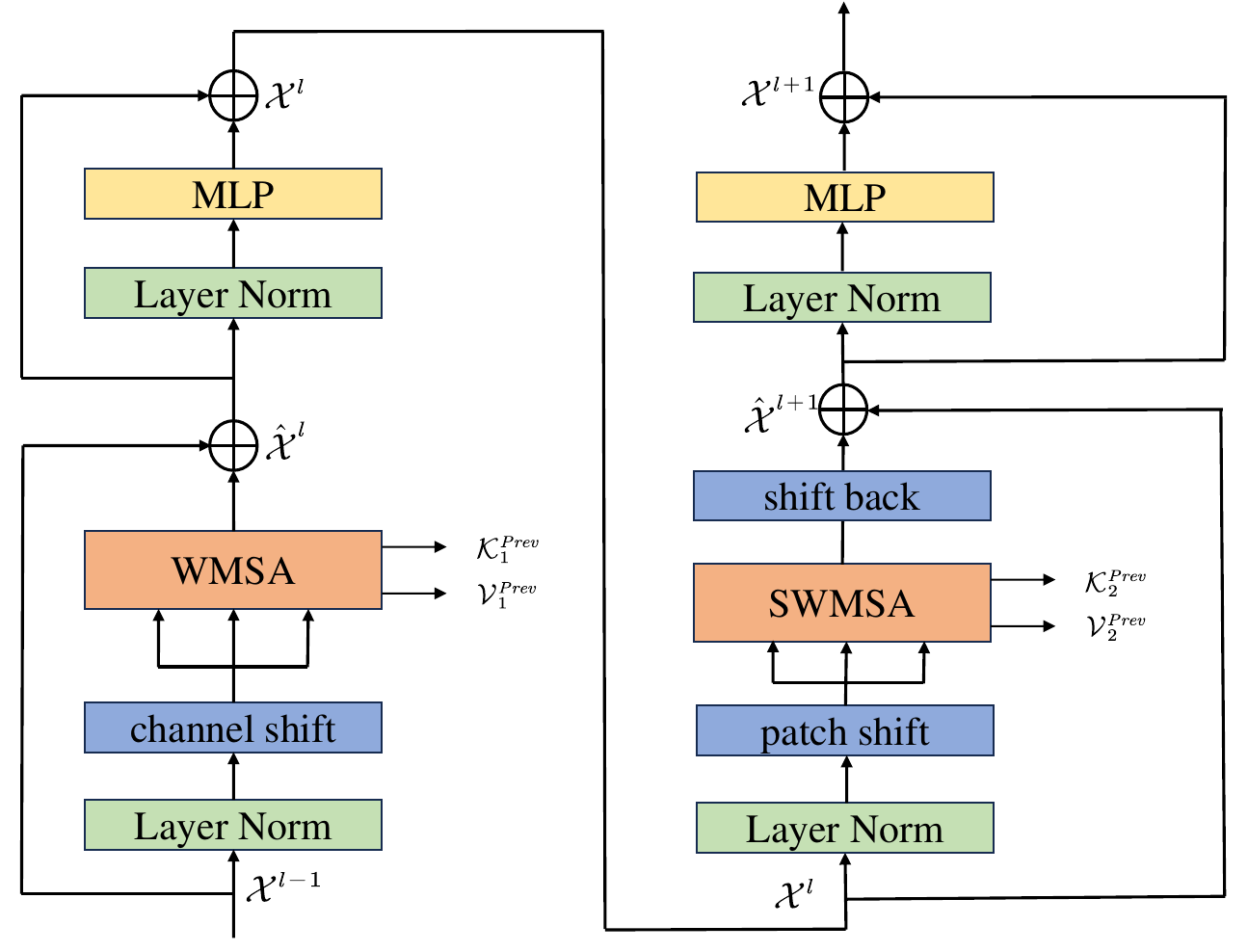}
\caption{Overview of two consecutive PatchShift 3D SwinTransformer blocks in the encoder.}
\label{figure8}
\end{figure}
The green, blue, red, yellow colors represent four successive frames. The features at different channel stamps in one frame are denoted as different shades of the same color. We can see from Fig. \ref{figure7} that the two shift methods perform shifting operations in orthogonal directions.
Previous work \cite{DBLP:journals/corr/abs-2207-13259} has revealed that patch shift and channel shift have a certain amount of complementary to each other, and the ability of temporal modeling can be enhanced by alternating them. We follow this idea and add channel shift to our model as a supplement for patch shift.

Fig. \ref{figure8} depicts two consecutive PatchShift 3D SwinTransformer blocks. Specifically, given the input feature \(\mathcal{X}^{l-1}\in \mathbb{R}^{T\times H\times W\times C}\) from the previous $(l-1)^{th}$ block, we first replace part of the channels of the current frame with neighboring frames following \cite{9008827}, which can be formulated as:
\begin{equation}\label{eq}
\mathcal{X}_{cs}^{l}=\text{ChannelShift}\left( \text{LN}\left( \mathcal{X}^{l-1} \right)\right) 
\end{equation}
where $\mathcal{X}_{cs}^{l}$ represents the output spatial-temporal mixed feature after channel shift, and $\text{LN}$ denotes layer normalization. The shift operation enables the integration of spatial information with temporal information in a zero-computation way, which reduces the computational complexity.

Then we set the size of each 3D window to $P_T\times P_H\times P_W$ and arrange the windows in a non-overlapping way to evenly split feature $\mathcal{X}_{cs}^{l}$ for efficient computation cost. Thus, feature $\mathcal{X}_{cs}^{l}$  is reshaped as $\mathcal{X}_{cs}^{l}\in \mathbb{R}^{N\times P\times C}$, where $N=\frac{THW}{P_TP_HP_W}$ denotes the number of windows, and $P=P_TP_HP_W$ represents the flattened window size. Afterwards we apply the window-based multi-head self attention (WMSA) module \cite{9710580}. Limiting attention computation in non-overlapping windows can bring the locality of convolution operations whiling saving computational resources. Finally we get the output $\mathcal{X}^l$ of the $l^{th}$ block through a feed forward network (FFN) and a shortcut connection like a standard transformer architecture. The process can be formulated as:
\begin{equation}\label{eq}
\hat{\mathcal{X}}^l=\text{WMSA}\left( \mathcal{X}_{cs}^{l} \right) +\mathcal{X}^{l-1}
\end{equation}
\begin{equation}\label{eq}
\mathcal{X}^l=\text{FFN}\left( \text{LN}\left( \hat{\mathcal{X}}^l \right) \right) +\hat{\mathcal{X}}^l
\end{equation}
where $\hat{\mathcal{X}}^l$ represents the output feature after WMSA. 
It should be noted that the $l^{th}$ block also outputs key feature $\mathcal{K}_{1}^{Prev}$ and value feature $\mathcal{V}_{1}^{Prev}$ from the WMSA module preparing for feature fusion in decoders.

After obtaining  $\mathcal{X}^l$ from the $l^{th}$ block, we first shift the patches of each frame along the temporal dimension with the specific pattern as:
\begin{equation}\label{eq}
\mathcal{X}_{ps}^{l+1}=\text{PatchShift}\left( \text{LN}\left( \mathcal{X}^l \right) \right) 
\end{equation}
where $\mathcal{X}_{ps}^{l+1}$ denotes the output feature after patch shift. Previous work \cite{DBLP:journals/corr/abs-2207-13259} has revealed that patch shift and channel shift are complementary to each other, so we adopt patch shift in the ${(l+1)}^{th}$ block and channel shift in the $l^{th}$ block in order to enhance the ability of temporal modeling.

Then we exploit the shifted window multi-head self-attention (SWMSA) module for cross-window connections. After the SWMSA, we follow \cite{DBLP:journals/corr/abs-2207-13259} and shift patches from different frames back to their original locations to keep the frame structure complete. Finally the output feature map $\mathcal{X}^{l+1}$ of the $(l+1)^{th}$ block is generated by a FFN and a shortcut connection. The process can be formulated as:
\begin{equation}\label{eq}
\hat{\mathcal{X}}_{ps}^{l+1}=\text{SWMSA}\left( \mathcal{X}_{ps}^{l+1} \right) 
\end{equation}
\begin{equation}\label{eq}
\hat{\mathcal{X}}^{l+1}=\text{ShiftBack}\left( \hat{\mathcal{X}}_{ps}^{l+1} \right) +\mathcal{X}^l
\end{equation}
\begin{equation}\label{eq}
\mathcal{X}^{l+1}=\text{FFN}\left( \text{LN}\left( \hat{\mathcal{X}}^{l+1} \right) \right) +\hat{\mathcal{X}}^{l+1}
\end{equation}
where $\hat{\mathcal{X}}_{ps}^{l+1}$ and $\hat{\mathcal{X}}^{l+1}$ denote the output features after the SWMSA and shift back, respectively. Note that the $(l+1)^{th}$ block also outputs the key feature $\mathcal{K}_{2}^{Prev}$ and value feature $\mathcal{V}_{2}^{Prev}$ preparing for feature fusion in decoders. By using shift operations, we extract the spatial-temporal feature in RF image sequences without high computation cost.

\subsubsection{Class Masking Attention Module (CMAM)}
\begin{figure}[htbp]
\centering
\includegraphics[scale=0.48]{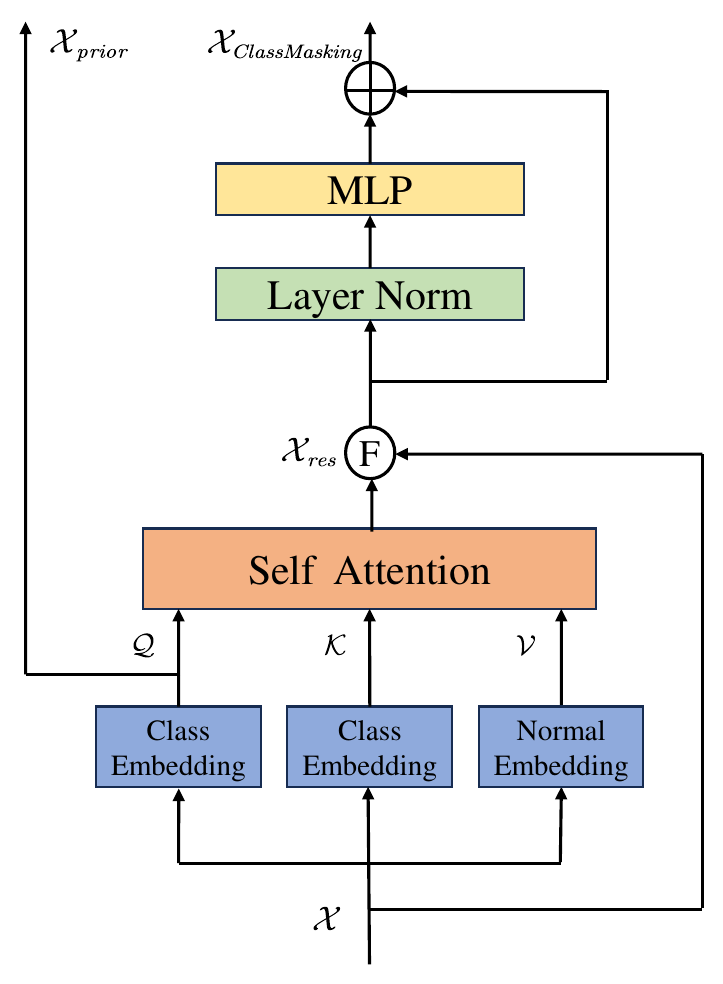}
\caption{Overview of CMAM in the encoder.}
\label{figure9}
\end{figure}

As illustrated in Fig. \ref{figure9}, the CMAM is proposed to capture the spatial-temporal contextual information from the perspective of the entire RF image sequence and generate enhanced feature maps with class-dependent semantic context information. It is designed to follow the PatchShift 3D SwinTransformer module at each stage of our model.

Given the output feature $\mathcal{X}\in \mathbb{R}^{T\times H\times W\times C}$ from previous 3D SwinTransformer module, we first utilize a class embedding layer to query feature $\mathcal{Q}$ and key feature $\mathcal{K}$. Class embedding layer is a linear layer that converts the channel $C$ to $class$, where $class$ denotes the number of categories. This operation means projecting the feature $\mathcal{X}$ into semantic space. 
\begin{equation}\label{eq1}
\mathcal{Q}=\text{ClassEmbedding}\left( \mathcal{X} \right) 
\end{equation}
\begin{equation}\label{eq1}
\mathcal{K}=\text{ClassEmbedding}\left( \mathcal{X} \right) 
\end{equation} 
where $\mathcal{Q}$ and $\mathcal{K}$ are both tensors of size $T\times H\times W\times class$. This strategy is inspired by some pioneering work \cite{Jain2021SeMaskSM} \cite{2018arXiv180308904Z}. Our intuition is: the output $\mathcal{Q}$ after class embedding layer contains class-dependent RF image semantic information to some extent, which can serve as the prior representation of current stage. Then $\mathcal{Q}$ is sent to auxiliary decoder for further ground truth supervision.

We also adopt a normal embedding layer to get value feature $\mathcal{V}$. Normal embedding layer is a linear layer which can convert the channel $C$ to the embedding dimension $C$, playing the same role as that in standard transformer architecture.
\begin{equation}\label{eq1}
\mathcal{V}=\text{NormalEmbedding}\left( \mathcal{X} \right) 
\end{equation}
where $\mathcal{V}$ is a tensor of size $T\times H\times W\times C$.

Next we perform reshape operations and calculate the similarities between $\mathcal{Q}$ and $\mathcal{K}$ :
\begin{equation}\label{eq}
\mathbf{Q} =\text{Reshape1}\left( \mathcal{Q} \right) 
\end{equation}
\begin{equation}\label{eq}
\mathbf{K} =\text{Reshape1} \left( \mathcal{K} \right) 
\end{equation}
\begin{equation}\label{eq}
\mathbf{S} =\text{Softmax} \left( \mathbf{Q}\otimes \mathbf{K} \right) 
\end{equation}
where Reshape1 is used to transform tensors $\mathcal{Q}$ and $\mathcal{K}$ to matrices $\mathbf{Q}$ and $\mathbf{K}$ of size $THW\times class$, respectively, $\otimes$ stands for matrix multiplication, and $\mathbf{S}$ is the similarity score between $\mathcal{Q}$ and $\mathcal{K}$. Considering that $\mathcal{Q}$ serves as the prior representation of current stage and is indirectly supervised by ground truth in auxiliary decoder, the score values in $\mathbf{S}$ contain semantic context information, which guide $\mathcal{V}$ to update as:
\begin{equation}\label{eq}
\mathbf{V} = \text{Reshape2} \left( \mathcal{V} \right) 
\end{equation}
\begin{equation}\label{eq}
\mathcal{R}=\text{Reshape3} \left(\mathbf{S} \otimes \mathbf{V} \right)
\end{equation}
where Reshape2 is used to make $\mathcal{V}$ a matrix of size $THW\times C$ and Reshape3 is used to make $\mathcal{R}$ a tensor of size $T\times H\times W\times C$.
 
To mitigate the vanishing gradient problem, we add a shortcut
connection and multiply $\mathcal{R}$ with a learnable scalar constant $\beta $ additionally for smooth finetuning. Finally we follow the traditional transformer architecture and get the final feature maps $\mathcal{X}_{ClassMasking}$ through a FFN and a shortcut connection. The process can be formulated as:
\begin{equation}\label{eq}
\mathcal{X}_{res}=\beta \mathcal{R}+\mathcal{X}
\end{equation}
\begin{equation}\label{eq}
\mathcal{X}_{ClassMasking}=\text{FFN}\left( \text{LN}\left( \mathcal{X}_{res} \right) \right) +\mathcal{X}_{res}
\end{equation}
where $\mathcal{X}_{res}$ is the output after attention operation.

\begin{figure}[htbp]
\centering
\includegraphics[scale=0.33]{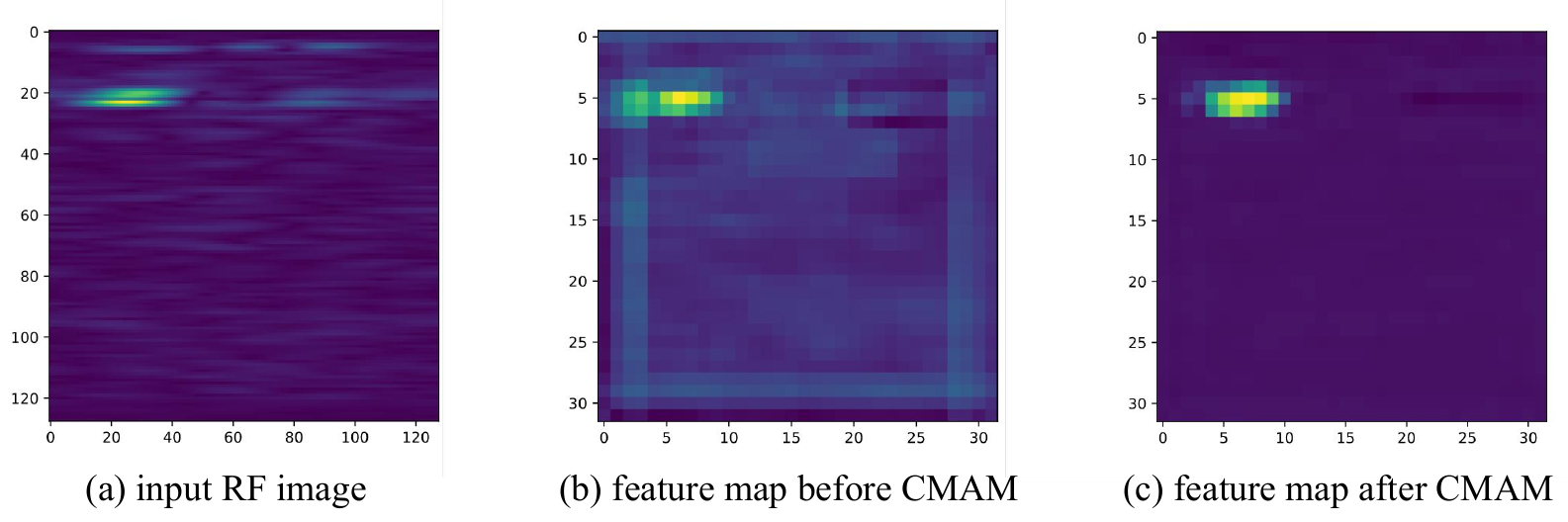}
\caption{Visualization of the feature maps before and after the CMAM module. The input RF image has the size of 128$\times$128. We select the feature maps with the size of 32$\times$32.}
\label{figure10}
\end{figure}

The output $\mathcal{X}_{ClassMasking}$ contains more spatial-temporal contextual information compared with the input $\mathcal{X}$. As shown in Fig.\ref{figure10}, the output feature map after the CMAM module focuses more on object areas. The results of the subsequent ablation experiments reveal that the CMAM enhances the performance of our model, particularly on average precision.

\subsection{Decoder}
In order to integrate the features and prior maps obtained from different encoder stages respectively, we utilize two different decoders.

\subsubsection{Main Decoder}
\begin{figure}[htbp]
\centering
\includegraphics[scale=0.36]{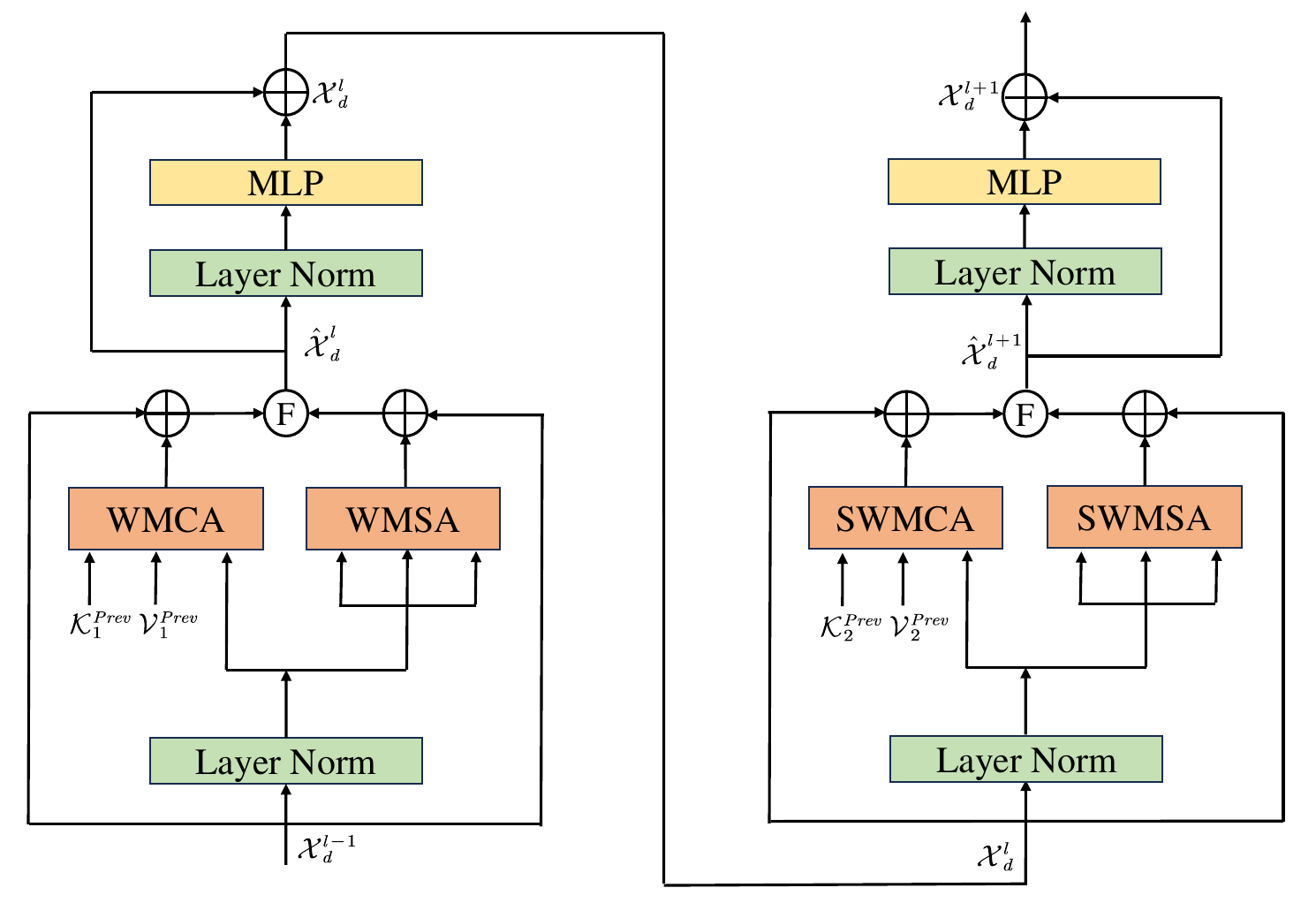}
\caption{Overview of two consecutive T-SwinTransformer blocks in the main decoder.}
\label{figure11}
\end{figure}
We employ the T-SwinTransformer blocks from \cite{9989400} as the main decoder in our work. Two consecutive T-SwinTransformer blocks are shown in Fig. \ref{figure11}. Inspired by the original transformer model \cite{2017Attention} in NLP, we believe that better prediction results can be obtained by integrating the feature from the encoder into the intrinsic feature of the decoder via cross attention operations.

Specifically, given the features $\mathcal{K}_{1}^{Prev}$, $\mathcal{V}_{1}^{Prev}$ from the corresponding PatchShift 3D SwinTransformer module and the inherent feature $\mathcal{X}_{d}^{l-1}$ from the previous $(l-1)^{th}$ decoder block, we first process them with the WMSA and the WMCA for cost-efficient computation:
\begin{equation}\label{eq}
\mathcal{SA}_l=\text{WMSA}\left( \text{LN}\left( \mathcal{X}_{d}^{l-1} \right) \right) +\mathcal{X}_{d}^{l-1}
\end{equation}
\begin{equation}\label{eq}
\mathcal{CA}_l=\text{WMCA}\left( \text{LN}\left( \mathcal{X}_{d}^{l-1} \right),\mathcal{K}_{1}^{Prev},\mathcal{V}_{1}^{Prev} \right) +\mathcal{X}_{d}^{l-1}
\end{equation}
where WMCA represents window based multi-head cross-attention using regular window partitioning configurations, $\mathcal{SA}_l$ denotes the inherent feature from the main decoder, and $\mathcal{CA}_l$ represents the feature obtained by interacting with the encoder. Then $\mathcal{SA}_l$ and $\mathcal{CA}_l$ are fused as:
\begin{equation}\label{eq}
\hat{\mathcal{X}}_{d}^l=\gamma\mathcal{CA}_l+\left( 1-\gamma \right)\mathcal{SA}_l 
\end{equation}
where $\gamma$ stands for a scaling factor that can be learned to compare the significance of the two outputs. After a FFN and a residual structure, the output feature map $\mathcal{X}_{d}^l$ of the $l^{th}$ block is obtained as:
\begin{equation}\label{eq}
\mathcal{X}_{d}^l=\text{FFN}\left( \text{LN}\left( \hat{\mathcal{X}}_{d}^l \right) \right) +\hat{\mathcal{X}}_{d}^l
\end{equation}
Next, we repeat the whole procedures above and shift windows during attention operation for cross-window connections:
\begin{equation}\label{eq}
\mathcal{SA}_{l+1}=\text{SWMSA}\left( \text{LN}\left( \mathcal{X}_{d}^l \right) \right) +\mathcal{X}_{d}^l
\end{equation}
\begin{equation}\label{eq}
\mathcal{CA}_{l+1}=\text{SWMCA}\left( \text{LN}\left( \mathcal{X}_{d}^l \right),\mathcal{K}_{2}^{Prev},\mathcal{V}_{2}^{Prev} \right) +\mathcal{X}_{d}^l
\end{equation}
\begin{equation}\label{eq}
\hat{\mathcal{X}}_{d}^{l+1}=\gamma \mathcal{CA}_{l+1}+\left( 1-\gamma \right)  \mathcal{SA}_{l+1} 
\end{equation}
\begin{equation}\label{eq}
\mathcal{X}_{d}^{l+1}=\text{FFN}\left( \text{LN}\left( \hat{\mathcal{X}}_{d}^{l+1} \right) \right) +\hat{\mathcal{X}}_{d}^{l+1}
\end{equation}
where SWMCA represents window based multi-head cross-attention using shifted window partitioning configurations.
Overall, the main decoder fuses encoder and decoder features in a learnable way in the T-SwinTransformer blocks.

\subsubsection{Auxiliary Decoder}
During the training stage, a lightweight FPN-like semantic decoder is used to provide ground truth supervision to the prior maps generated from the CMAM at each stage of the encoder. Considering that all prior maps from different stage have the same channel dimension of $class$, we aggregate them only with some upsampling and summation operations. More details are displayed in Fig. \ref{figure3}.

\subsection{Loss Function}
In this paper, we utilize the auxiliary loss function to supervise the training of our proposed method. After passing through the network, confidence maps (ConfMaps) $\hat{\mathcal{C}}\in \mathbb{R}^{T\times H\times W\times class}$ and prior maps $
\hat{\mathcal{P}}\in \mathbb{R}^{T\times H\times W\times class}$ are predicted from main decoder and auxiliary decoder, respectively. We use binary cross entropy loss to supervise the output of the main decoder, and the main loss function is defined as:
\begin{equation}\label{eq26}
\begin{aligned}
  l_{main}=&-\sum_{class}\sum_{i,j}\left\{\mathcal{GT}_{i,j}^{class}\log \hat{\mathcal{C}}_{i,j}^{class} \right.\\
           &\left.+\left( 1-\mathcal{GT}_{i,j}^{class} \right) \log \left( 1-\hat{\mathcal{C}}_{i,j}^{class} \right) \right\}
\end{aligned}
\end{equation}

\noindent where $\mathcal{GT}_{i,j}^{class}$ indicates the ground truth generated by CRF at coordinate $(i,j)$ for category label $class$, and $\hat{\mathcal{C}}_{i,j}^{class}$ indicates the predictions generated by the main decoder at coordinate $(i,j)$ for category label $class$.

Then we add a specific auxiliary loss function to supervise the output of auxiliary decoder. The auxiliary loss function is consistent with the main loss function, which can be defined as:
\begin{equation}\label{eq27}
\begin{aligned}
  l_{aux}=&-\sum_{class}\sum_{i,j}\left\{\mathcal{GT}_{i,j}^{class}\log \hat{\mathcal{P}}_{i,j}^{class} \right.\\
           &\left.+\left( 1-\mathcal{GT}_{i,j}^{class} \right) \log \left( 1-\hat{\mathcal{P}}_{i,j}^{class} \right) \right\}
\end{aligned}
\end{equation}
\noindent where $\hat{\mathcal{P}}_{i,j}^{class}$ indicates the prior maps generated by the auxiliary decoder at coordinate $(i,j)$ for category label $class$.

Furthermore, we use the parameter $\alpha$ to balance the weight of the main loss and auxiliary loss, i.e.,
\begin{equation}\label{eq28}
l=l_{main}+\alpha l_{aux}
\end{equation}
It should be noted that we only use the auxiliary loss in the training phase.

\section{Experiments}
In this section, we present the experimental evaluation of our model. Firstly we describe the dataset and the evaluation metric that we utilize. Then we give details concerning the experiments. Next we compare our model with the SOTA and analyse the results quantitatively and qualitatively. Finally we perform some ablation studies of Mask-RadarNet.

\subsection{Dataset}
We train our Mask-RadarNet with the training data of the CRUW dataset\cite{9353210}. The CRUW dataset contains 3.5 hours with 30 FPS of camera-radar data in different driving scenarios, which are collected with an RGB camera and 77 GHz FMCW MMW radar antenna arrays. The high frame rate makes the CRUW dataset being appropriate for evaluating the temporal models. Data in this dataset are processed and presented as Range-Azimuth (RA) heatmaps, depicting a bird-eye-view of the scene seen from the ego-vehicle. RA heatmaps can be described as images with a resolution of 128×128 and the intensity depicts the magnitude of the RF signal. The cross-modal supervision framework in \cite{9353210} labels the collected objects with camera-radar locators, which makes full use of FMCW radar and offers an appropriate capability for range estimation free from any systematic bias. Generally, there are around $2.6\times 10^5$ objects in the CRUW dataset, 92 percent of which are utilized for training and 8 percent for testing. Besides, this dataset contains some vision-fail scenarios, allowing the model to be tested in extreme environments.

\subsection{Evaluation Metrics}
To evaluate model on the CRUW dataset, a new metric called object location similarity (OLS) \cite{9353210} is defined. This method depicts the correlation between two point-based detections while taking into account their distance, classes, and scale characteristics. Specifically, the OLS can be written as:
\begin{equation}\label{eq1}
\text{OLS}=\exp \left\{ \frac{-d^2}{2\left( S\cdot K_{cls} \right) ^2} \right\} 
\end{equation}

\noindent where \emph{d} is the distance between the two points in an RF image; \emph{S} is the object distance from sensors, indicating
object scale information; and $K_{cls}$ is a per-class constant representing the error tolerance for class \emph{cls}, which can be determined by the object average size of the corresponding class.

The evaluation approach employed in our work is consistent with that of \cite{9989400}. Here is a synopsis of the procedure:

1) Get all the 8-neighbor peaks in all $class$ channels in ConfMaps within the 3 × 3 window as a peak set $P$.

2) Pick the peak $p^*\in P$ with the highest confidence score, and remove it from the set $P$ to the final peak set $P^*$.

3) Calculate the OLS with each of the rest peaks $p_i\in P$. If the OLS between $p^*$ and $p_i$ is greater than a threshold, we remove $p_i$ from the set $P$.

4) Repeat steps 2 and 3 until the set $P$ becomes empty.
\begin{table*}
\begin{center}
\caption{Results of different models on the CRUW dataset.}
\label{tab1}
\setlength{\tabcolsep}{6pt}
\renewcommand{\arraystretch}{1.5}
\begin{tabular}{cccccccccccc}
\toprule
\multirow{2}{*}{Model} & \multicolumn{2}{c}{All}                      & \multicolumn{2}{c}{Pedestrian}               & \multicolumn{2}{c}{Cyclist}                  & \multicolumn{2}{c}{Car}                      & \multicolumn{1}{l}{\multirow{2}{*}{GFLOPs}} & \multicolumn{1}{l}{\multirow{2}{*}{Parameter(M)}} & \multicolumn{1}{l}{\multirow{2}{*}{Time(H)}} \\ 
\cmidrule(r){2-3}
\cmidrule(r){4-5}
\cmidrule(r){6-7}
\cmidrule(r){8-9}

                                & AP(\%) & \multicolumn{1}{l}{AR(\%)} & AP(\%) & \multicolumn{1}{l}{AR(\%)} & AP(\%) & \multicolumn{1}{l}{AR(\%)} & AP(\%) & \multicolumn{1}{l}{AR(\%)} & \multicolumn{1}{l}{}                                 & \multicolumn{1}{l}{}                                       \\ \midrule
RODNet-CDC\cite{9353210}             & 76.33           & 79.28                               & 77.11           & 79.64                               & 69.39           & 70.02                               & 82.91          & 89.13                               & 280.03                                               & 34.52
& \textbf{9}
\\
RODNet-HG\cite{9353210}              &79.43           & 83.59                              & 78.90           & 83.81                              & 76.69         & 78.85                               & 83.36          & 88.55                               & 2144.86                                              & 129.19
& 65
\\ 
RODNet-HGWI\cite{9353210}            & 78.06           & 81.07                              & 79.47         & 81.85                               & 70.35           & 71.40                               & 84.39           & 90.05                             & 5949.68                                              & 61.29  &330                                                    \\
DCSN\cite{10.1145/3460426.3463657}                   & 75.30           & 79.92                               & 76.70           & 81.50                              & 66.78           & 69.04                              & 82.56           & 89.52                              & 3039.89                                              & \textbf{28.10} &204                                             \\
T-RODNet\cite{9989400}               & 83.27           & 86.98                              & 82.19           & 85.41                               & 82.28          & 84.30                   & \textbf{86.22}           & \textbf{92.53}                      & 182.53                                               & 44.31 &15                                                    \\
SS-RODNet\cite{bu8}               & 83.07           & 86.43                              & 81.37           & 84.61                               & 83.34           & 85.11                      & 85.55           & 90.86                      & \textbf{172.80}                                               & 33.10 &15                                                    \\
Mask-RadarNet (Ours)              & \textbf{84.29}  & \textbf{87.36}                      & \textbf{82.74}  & \textbf{85.80}                      & \textbf{85.06}           & \textbf{86.67}                               & 85.96           & 90.66                               & 176.91                                      & 32.12  &14\\
\bottomrule
\end{tabular}
\end{center}
\end{table*}

The whole procedure is similar to previous work for pose estimation\cite{Lin2014MicrosoftCC}. Finally, average precision (AP) and average recall (AR) are calculated through the variation of OLS threshold between 0.5 to 0.9 with steps of 0.05. The AP and AR are the main evaluation metrics for object detection.

\subsection{Implementation Details}
We run all experiments on Python 3.8, PyTorch 1.10.1 and Ubuntu 18.04. All training procedures have been performed on RTX 3080 GPU. With the purpose of comparing the performance of every model, we split the 40 sequences in the CRUW dataset into two parts: 36 sequences for training and 4 sequences for testing, which is the same in \cite{9989400}. We input 16 consecutive RF frames at one time and treat the real and imaginary values as two channels in an RF image. So the input size is 2 × 16 × 128 × 128. For the encoder, we set the size of convolutional kernels as 9 × 5 × 5 for 3D Convolutional Embedding and 3D Convolutional DownSampling. The work in \cite{Ding2022ScalingUY} demonstrates that employing the larger convolutional kernel rather than a series of small kernels might be a better option. For the PatchShift 3D SwinTransformer module, we carry out patch shift with Pattern C in Fig. \ref{figure5} and apply channel shift with a shift ratio of 0.25, identical to that described in \cite{9008827}. Furthermore, we use a 3D window of 4 × 4 × 4 for window partition. The number of heads of multi-head self-attention used in different encoder stages is [2, 4, 8]. The corresponding part in the main decoder has the same settings. For the loss function, we empirically set the auxiliary loss weight $\alpha$ as 0.4. The Adam optimizer is utilized for optimization, with the starting learning rate set to 0.0001.
\begin{figure*}[htbp]
\centering
\includegraphics[scale=0.458]{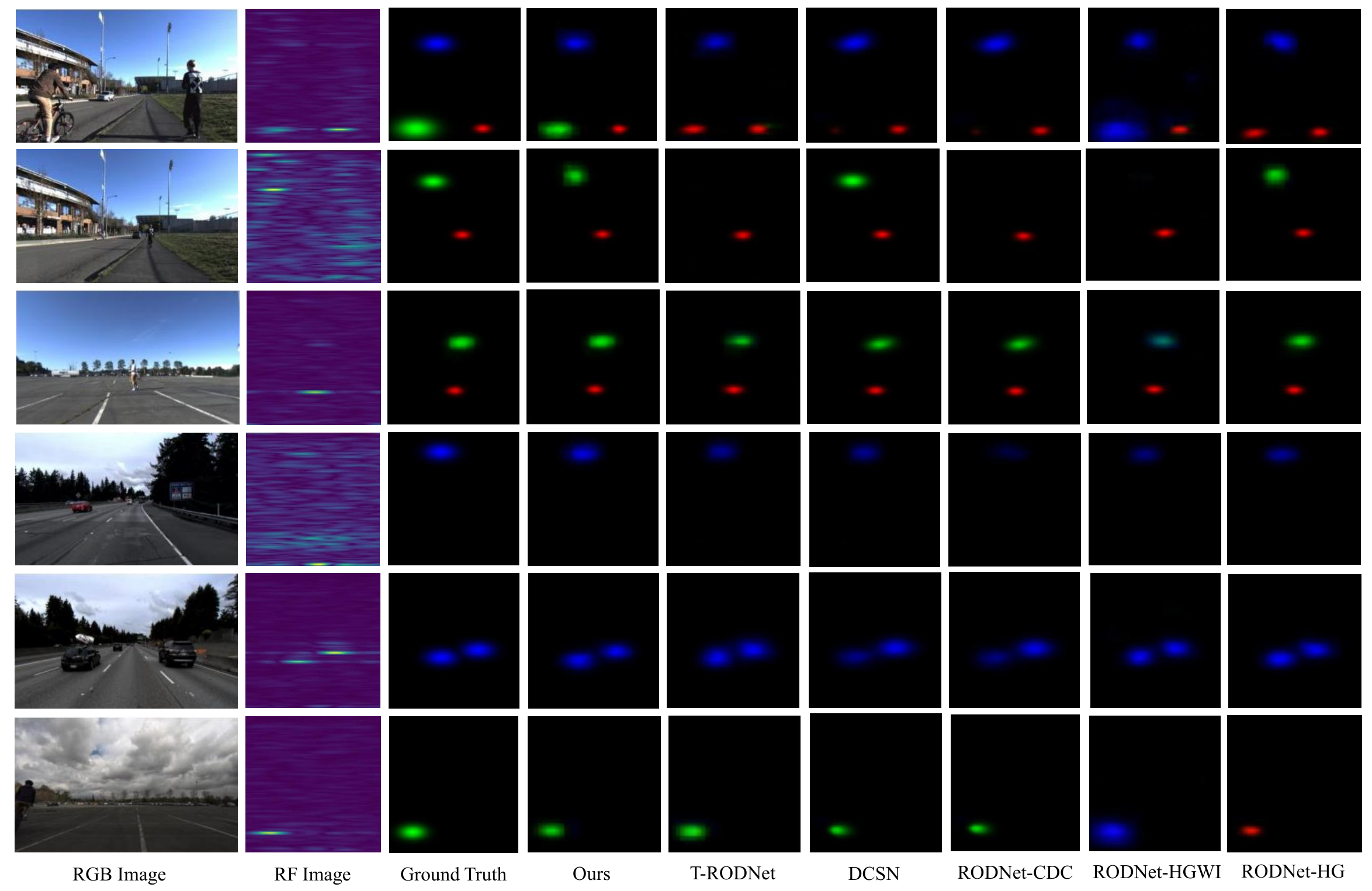}
\caption{Visual comparison with other models on the CRUW dataset. Mask-RadarNet outperforms others in all scenarios. The colors red, blue, and green correspond to different categories of objects: pedestrian, car, and cyclist, respectively.}
\label{figure13}
\end{figure*}

\begin{table*}
\begin{center}
\caption{Results of ablation experiments for patch shift.}
\label{tab1}
\setlength{\tabcolsep}{5mm}
\renewcommand{\arraystretch}{1.5}
\begin{tabular}{ccccccccc}
\toprule
\multirow{2}{*}{PatchShift} & \multicolumn{2}{c}{All}                      & \multicolumn{2}{c}{Pedestrian}               & \multicolumn{2}{c}{Cyclist}                  & \multicolumn{2}{c}{Car}                      \\
\cmidrule(r){2-3}
\cmidrule(r){4-5}
\cmidrule(r){6-7}
\cmidrule(r){8-9}
                                     & AP(\%) & \multicolumn{1}{l}{AR(\%)} & AP(\%) & \multicolumn{1}{l}{AR(\%)} & AP(\%) & \multicolumn{1}{l}{AR(\%)} & AP(\%) & \multicolumn{1}{l}{AR(\%)} \\
\hline
No Shift               & 81.35           & 85.44                               & 78.42           & 82.29                               & 82.67           & 84.41                               & 84.61           & \textbf{91.72}                      \\
Pattern A                   & 82.06           & 85.83                               & 81.00           & 83.75                               & 81.97           & 85.02                               & 83.87           & 90.10                               \\
Pattern B                   & 82.20           & 85.97                               & 82.25           & \textbf{86.08}                      & 79.61           & 80.92                               & 85.01           & 91.43                               \\
Pattern C                   & \textbf{84.29}  & \textbf{87.36}                      & \textbf{82.74}  & 85.80                               & \textbf{85.06}  & \textbf{86.67}                      & \textbf{85.96}  & 90.66     \\               
\bottomrule
\end{tabular}
\end{center}
\end{table*}

\begin{table}
\begin{center}
\caption{Results of ablation experiments for CMAM.}
\label{tab1}
\setlength{\tabcolsep}{6mm}
\renewcommand{\arraystretch}{1.5}
\begin{tabular}{clcc}
\toprule
\hline
\multicolumn{2}{c}{Module}                         & AP(\%) & AR(\%) \\ \hline
\multicolumn{2}{c}{None}                           & 81.69      & 85.92      \\ 
\multicolumn{2}{c}{CMAM}                           & \textbf{84.29}      & \textbf{87.36}      \\ 
\multicolumn{2}{c}{Traditional Transformer Module} & 81.72      & 85.59      \\ \hline 
\end{tabular}
\end{center}
\end{table}

\begin{table}
\begin{center}
\caption{Results of different auxiliary loss weight.}
\label{tab1}
\setlength{\tabcolsep}{7.1mm}
\renewcommand{\arraystretch}{1.5}
\begin{tabular}{clcc}
\toprule
\hline
\multicolumn{2}{c}{Auxiliary Loss Weight $\alpha$} & AP(\%) & AR(\%) \\ \hline
\multicolumn{2}{c}{$\alpha$ = 0}              & 81.71      & 85.18      \\
\multicolumn{2}{c}{$\alpha$ = 0.1}              & 82.34      & 86.13      \\
\multicolumn{2}{c}{$\alpha$ = 0.2}              & 82.94      & 86.00      \\
\multicolumn{2}{c}{$\alpha$ = 0.3}              & 82.89      & 86.82      \\
\multicolumn{2}{c}{$\alpha$ = 0.4}              & \textbf{84.29}      & 87.36      \\
\multicolumn{2}{c}{$\alpha$ = 0.5}              & 83.65      & 87.11      \\
\multicolumn{2}{c}{$\alpha$ = 0.6}              & 83.44      & \textbf{87.53}      \\
\multicolumn{2}{c}{$\alpha$ = 0.7}              & 82.73      & 86.47      \\
\multicolumn{2}{c}{$\alpha$ = 0.8}              & 82.32      & 86.23      \\
\multicolumn{2}{c}{$\alpha$ = 0.9}              & 82.36      & 87.03     \\ \hline 
\end{tabular}
\end{center}
\end{table}

\subsection{Comparisons with SOTA}
We perform some experimental analysis of several previous algorithms on the CRUW dataset, including the SOTA T-RODNet\cite{9989400}. All models are tested under the same conditions without using any data enhancement or pretrained models. Aside from AP and AR, we compare models in terms of model efficiency, where we adopt two efficiency-related evaluation metrics, i.e., the size of model parameters and GFLOPs. We also document the time required for model training.

\subsubsection{Quantitative Results}
To facilitate a qualitative comparison, the numerical results on the CRUW dataset are presented in Table \uppercase\expandafter{\romannumeral1}, revealing that the Mask-RadarNet outperforms other models in general. The AP and AR of Mask-RadarNet in all categories are 1.02$\%$ and 0.38$\%$, respectively, higher than the T-RODNet (SOTA). Notably, Mask-RadarNet significantly improves detection performance on small objects such as pedestrians and cyclists.
In addition, its GFLOPs and parameters are less than those of the T-RODNet. 
It means that the Mask-RadarNet achieves stronger detection results with lower computational complexity. Compared to previous convolution-based temporal modeling methods utilized in radar object detection, such as temporal deformable convolution network in \cite{9353210} and dimensional apart module in \cite{9989400}, the shift operation we employ can significantly reduce the computation cost. As a result, our model exhibits lower computational complexity. In addition, the reason for stronger detection results lies in the fact that the CMAM we design is capable of merging semantic spatial-temporal context in the feature map acquired by the encoder and generating semantic prior maps. Simultaneously, our auxiliary decoder is capable of aggregating the semantic prior maps generated by the CMAM at various stages. The final semantic prior map, produced after aggregation, is supervised by the ground truth, resulting in the auxiliary loss. Our subsequent ablation experiments demonstrate that the model performance can be enhanced by incorporating auxiliary loss during training.

\subsubsection{Qualitative Results}
To make a qualitative comparison, Fig. \ref{figure13} shows some visual examples of different models. It is evident that the Mask-RadarNet is able to generate better predictions than other methods. We can observe that the Mask-RadarNet predicts the location and the category of objects accurately, while some other models can predict exact location but mix up categories, such as (1st row, 5th col), (1st row, 8th col) and (6th row, 9th col) in Fig. \ref{figure13}. Our Mask-RadarNet has better ability of class-dependent semantic feature modeling, thus improving detection performance. Overall, the predictions from our Mask-RadarNet resemble the ground truth for the data.

\subsection{Ablation Studies}
In this section, we perform some ablation studies of our Mask-RadarNet on the CRUW dataset. It should be noted that all experiments are conducted under identical conditions to achieve fair comparisons.

\subsubsection{The effectiveness of Patch Shift}
To evaluate patch shift operation, we conduct experiments with different shift pattern settings. Note that all experiments in this section adopt the CMAM and set the auxiliary loss weight $\alpha$ as 0.4. We only change patch shift patterns and remain other components unchanged.
First we remove all shift operations. Table \uppercase\expandafter{\romannumeral2} indicates that the Mask-RadarNet without shift operation performs worse than any Mask-RadarNet with shift operation, which demonstrates significance of patch shift. In other words, the Mask-RadarNet profits from the shift operation to efficiently learn the spatial-temporal feature. Then we test different kinds of patch shift patterns in Fig. \ref{figure5}. Pattern A only shifts patches within 3 neighboring frames, while Pattern B has a temporal of 4 and Pattern C has a temporal field of 9. Table \uppercase\expandafter{\romannumeral2} shows that the performance of Mask-RadarNet steadily improves as the temporal field expands. The AP and AR for cyclist improve greatly with Pattern C. This may be because the change of cyclist in RF images is the most evident in the temporal field of 9 compared with pedestrian and car. In conclusion, our patch shift operations can better capture features and improve detection performance, and we apply Pattern C for patch shift in our model.

\subsubsection{The effectiveness of CMAM}
We conduct three groups of experiments to explore the impact of CMAM on our model. In the first group, we remove the CMAM module while remaining other components unchanged. In the second group, we keep the CMAM module. In the last group, the CMAM module is replaced with the traditional transformer encoder module \cite{Dosovitskiy2020AnII}. The experimental results are presented in Table \uppercase\expandafter{\romannumeral3}. To facilitate comparisons, we utilize the AP and AR of all categories as indicators. It is observed that the AP and AR obtained by the Mask-RadarNet without the CMAM decrease remarkably. And the detection results have not been significantly improved when the CMAM module is substituted by the standard transformer encoder module. This reveals that the simple attention operations cannot provide the effectiveness of the CMAM module, and the CMAM module can better capture the spatial-temporal semantic context.

\subsubsection{The effectiveness of auxiliary loss weight}
It is essential to set an appropriate loss weight in the auxiliary decoder, so that the model can regard the spatial-temporal semantic context as a supplemental signal rather than the main prediction. We conduct several experiments with setting the auxiliary loss weight $\alpha$ between 0 and 1. The numerical results are shown in Table \uppercase\expandafter{\romannumeral4}. To facilitate comparisons, we utilize the AP and AR of all categories as indicators. In general, the introduced auxiliary loss aids in optimizing the training process while not influencing the inference process. We can comprehend auxiliary loss from the perspective of gradient descent. When the model incorporates an auxiliary branch, the gradient of the parameters in the model originates from both the main branch and the auxiliary branch. The auxiliary loss utilized in our Mask-RadarNet is aligned with the main loss, ensuring that the gradient it provides aligns with the direction of the main branch, which benefits the training of the main branch. We can observe from Table \uppercase\expandafter{\romannumeral4} that $\alpha = 0.4$ yields the best performance which achieves a good balance between modeling feature context and semantic context in RF image sequence.

\section{Conclusion}
In this paper, we proposed a transformer-based model called Mask-RadarNet for radar object detection, which addresses the issue of leaving out spatial-temporal semantic context during encoding. Specifically, our Mask-RadarNet exploits the combination of interleaved convolution and attention operations on
multiframe RF images to extract both local and global features. Patch shift was introduced to attention operations as an efficient method for temporal modeling. Moreover, we designed a simple yet effective module called CMAM to capture the spatial-temporal contextual information in our encoder. Besides, a lightweight auxiliary decoder was proposed to aggregate prior maps that guide the model in fine-tuning the features it captures, thereby enhancing the network's performance. The experimental results showed that our Mask-RadarNet achieves SOTA performance with lower GFLOPs and fewer parameters on the CRUW dataset.

\nocite{*}
\bibliographystyle{IEEEtran}
\bibliography{main}

\end{document}